\documentclass[11pt,a4paper]{article}
\usepackage{CJKutf8}

\usepackage[hyperref]{acl2020}
\usepackage{times}
\usepackage{latexsym}
\usepackage{graphicx}
\usepackage{caption}
\usepackage{multirow}

\usepackage{microtype}

\aclfinalcopy 
\def\aclpaperid{465} 

\title{Lexically Constrained Neural Machine Translation \\with Levenshtein Transformer}

\author{Raymond Hendy Susanto, Shamil Chollampatt, \and Liling Tan\\
  Rakuten Institute of Technology Singapore \\ Rakuten, Inc. \\
  \texttt{\{first.last\}@rakuten.com} \\}
\date{}

\begin{document}
\maketitle
\begin{abstract}
This paper proposes a simple and effective algorithm for incorporating lexical constraints in neural machine translation. Previous work either required re-training existing models with the lexical constraints or incorporating them during beam search decoding with significantly higher computational overheads. Leveraging the flexibility and speed of a recently proposed Levenshtein Transformer model \cite{gu2019levenshtein}, our method injects terminology constraints at inference time without any impact on decoding speed. Our method does not require any modification to the training procedure and can be easily applied at runtime with custom dictionaries. Experiments on English-German WMT datasets show that our approach improves an unconstrained baseline and previous approaches.
\end{abstract}

\section{Introduction}

Neural machine translation (NMT) systems can generate higher-quality translations than phrase-based MT systems, but they come at the cost of losing control over how translations are generated. Without the explicit link between the source and the target vocabulary, enforcing specific terminological translation in domain-specific settings becomes painfully difficult for NMT systems. Consider an example where we have a Chinese-English NMT system trained for the E-commerce domain, and there is no prior knowledge of the brand name ``\begin{CJK*}{UTF8}{gbsn}红米\end{CJK*}'' in the training data, the system would translate the input term literally as ``\textit{red} (\begin{CJK*}{UTF8}{gbsn}红\end{CJK*}) \textit{rice} (\begin{CJK*}{UTF8}{gbsn}米\end{CJK*})'' instead of ``\textit{Redmi}''. In such scenarios, machine translation users often maintain in-domain dictionaries to ensure that specific information is translated accurately and consistently.

A line of previous work that tried to address this problem required re-training the NMT models with lexical constraints, either by a placeholder mechanism \cite{crego-2016} or via code-mixed training \cite{song-etal-2019-code,dinu-etal-2019-training}. However, they do not reliably guarantee the presence of the constraints at test time. Another approach focused on constrained beam search decoding \cite{hokamp-liu-2017-lexically,post-vilar-2018-fast,hu-etal-2019-improved}. Although the latter approach has higher control over the target constraint terms, they significantly slow down the decoding.

Different from the existing line of work, we invoke lexical constraints using a non-autoregressive approach.\footnote{In literature, non-autoregressive NMT decoding mostly refers to those that do not generate tokens sequentially, although they perform iterative refinement \cite{Lee:18}.} To do this, we use Levenshtein Transformer (LevT) \cite{gu2019levenshtein}, an edit-based generation model that performs deletion and insertion operations during inference iteratively. LevT achieves substantially higher inference speed compared to beam search without affecting quality. We add a \textit{constraint insertion} step in LevT decoding to seamlessly decode the target language sequence while adhering to specific lexical constraints, achieving the same speed as standard LevT decoding. 

\section{Related Work}

Previous approaches integrated lexical constraints in NMT either via constrained training or decoding. \citet{crego-2016} replaced entities with placeholders that remained unchanged during translation and placed them back in a post-processing step. \citet{song-etal-2019-code} trained a Transformer \cite{vaswani2017} model by augmenting the data to include the constraint target phrases in the source sentence. \citet{dinu-etal-2019-training} proposed a similar idea and additionally used factored training. Other approaches proposed enforcement of lexical constraints during inference with various improvements to constraint-aware beam search, such as grid beam search \cite{hokamp-liu-2017-lexically}, dynamic beam allocation \cite{post-vilar-2018-fast}, and its optimized vectorized version \cite{hu-etal-2019-improved}. \citet{hasler-etal-2018-neural} built finite-state acceptors to integrate constraints in a multi-stack decoder. These lexically-constrained decoding approaches rely on autoregressive inference that generates one target token at a time, which makes it difficult to parallelize the decoder and monotonically increases decoding time. While being mostly effective at forcing the inclusion of pre-specified terms in the output, these approaches further slow down the beam search process. \citet{post-vilar-2018-fast} reported 3$\times$ slow down compared to standard beam search.

Non-autoregressive neural machine translation (NAT) \cite{gu2017non} attempts to move away from the conventional autoregressive decoding. Such a direction enables parallelization during sequence generation that results in lower inference latency. 
Recent NAT approaches treat inference as an iterative refinement process, first proposed by \citet{Lee:18}. 
Following this direction, it is intuitive to perform decoding using ``edit'' operations, such as insertion \cite{stern2019ins} or both insertion and deletion (LevT, \citet{gu2019levenshtein}). The LevT model has been shown to outperform existing refinement-based models, such as \citet{ghazvininejad2019MaskPredict} and performs comparably to autoregressive Transformer models. Our method integrates lexical constraints in NAT decoding utilizing the flexibility, speed, and performance of LevT.

\section{Levenshtein Transformer}

Levenshtein Transformer (LevT) \cite{gu2019levenshtein} has an encoder-decoder framework based on Transformer architecture \cite{vaswani2017} with multi-headed self-attention and feed-forward networks. Unlike token generation in a typical Transformer model, LevT decoder models a Markov Decision Process (MDP) that iteratively refines the generated tokens by alternating between the insertion and deletion operations. After embedding the source input through a Transformer encoder block, the LevT decoder follows the MDP formulation for each sequence at the $k$-th iteration $\mathbf{y}^k = (y_1, y_2, ..., y_n)$, where $y_1$ and $y_n$ are the start (\texttt{<s>}) and end (\texttt{</s>}) symbols. The decoder then generates $\mathbf{y}^{k+1}$ by performing deletion and insertion operations via three classifiers that run sequentially:

\begin{enumerate}
    \item \textbf{Deletion Classifier}, which predicts for each token position whether they should be ``kept'' or ``deleted'',
    \item \textbf{Placeholder Classifier}, which predicts the number of tokens to be inserted between every two consecutive tokens and then inserts the corresponding number of placeholder \texttt{[PLH]} tokens,
    \item \textbf{Token Classifier}, which predicts for each \texttt{[PLH]} token an actual target token.
\end{enumerate}
Each prediction is conditioned on the source text and the current target text. The same Transformer decoder block is shared among the three classifiers. Decoding stops when the current target text does not change, or a maximum number of refinement iterations has been reached.

The LevT model is trained using sequence-level knowledge distillation \cite{kim-rush-2016-sequence} from a Transformer teacher whose beam search output is used as ground truth during training. We refer the readers to \cite{gu2019levenshtein} for a detailed description of the LevT model and training routine.

\begin{figure}
\centering
\includegraphics[width=0.47\textwidth]{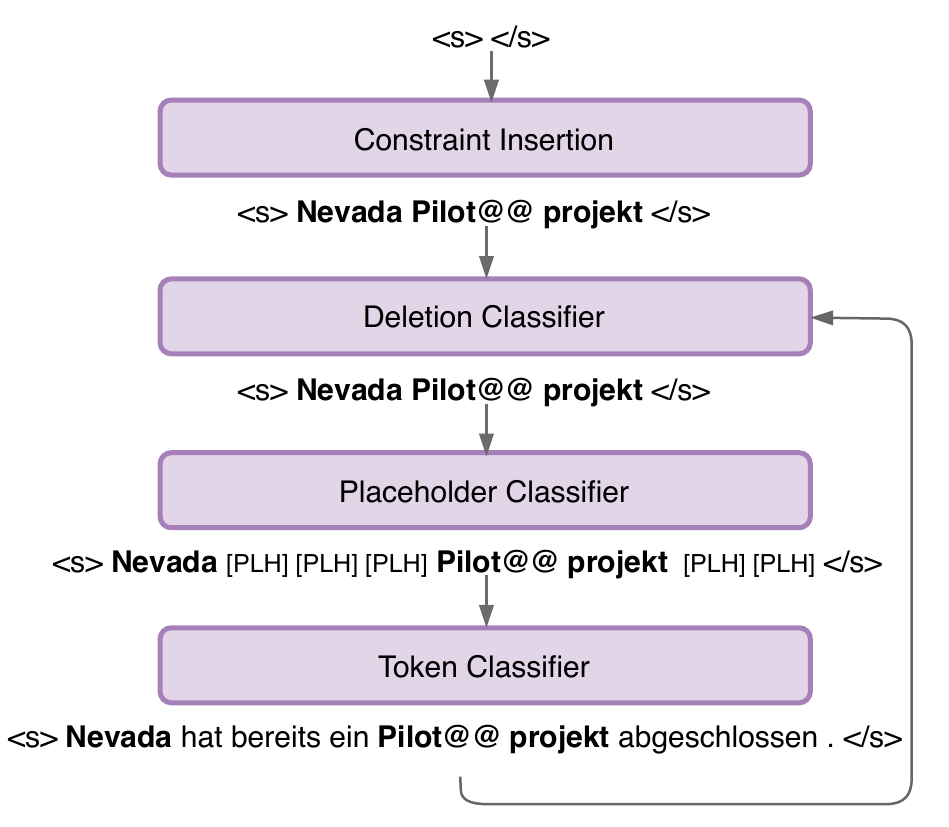}
\caption{Levenshtein Transformer decoding with lexical constraints for English-German MT. The source sentence is \textit{Nevada has completed a pilot project.} and the target constraints are [\textit{Nevada}, \textit{Pilot@@ projekt}]. Encoder and attention components are not shown.}
\label{fig:levt_constraint}
\end{figure}

\section{Incorporating Lexical Constraints}

For sequence generation, the LevT decoder typically starts the first iteration of the decoding process with only the sentence boundary tokens $\mathbf{y}^0=\texttt{<s>}\texttt{</s>}$. To incorporate lexical constraints, we populate the $\mathbf{y}^0$ sequence before the first deletion operation with the target constraints, as shown in Figure~\ref{fig:levt_constraint}. The initial target sequence will pass through the deletion, placeholder, and insertion classifiers sequentially, and the modified sequence will be refined for several iterations. The decoding steps are explained in detail below.

\paragraph{Constraint Insertion} More formally, given a list of $m$ target constraints $C_1, C_2, ..., C_m$, where each constraint $C_i$ is possibly a multi-token phrase $C_i = w^i_1,  w^i_2, ... , w^i_{|C_i|}$, we insert the constraints into the decoding sequence before the deletion operation to form $\mathbf{y}^0 = \texttt{<s>} \, C_1 \, C_2 \, ... \, C_n \texttt{</s>} $.

\paragraph{Deletion Operation}  Next, $\mathbf{y}^0$ passes through the \textit{deletion classifier} to decide which  $w^i_j$ token to remove. If the deletion operation is allowed on the constraint tokens, the presence of each constraint in the final output is not guaranteed, especially when the supplied constraints are out of context for the decoder. To mitigate this problem, we optionally disallow the deletion operation on the constraint tokens by introducing a \textit{constraint mask} to indicate the positions of constraint tokens in the sequence. We forcefully set the deletion classifier prediction for all positions in this mask to ``keep''. The positions in this mask are re-computed accordingly after each deletion and insertion operation. 

\paragraph{Insertion Operation} Finally, the $\mathbf{y}^0$ passes through the \textit{placeholder classifier} to predict the number of tokens to be inserted and generate the corresponding number of \texttt{[PLH]} tokens and the \textit{token classifier} assigns an actual target token for every \texttt{[PLH]} token. Each constraint may contain multiple tokens, and the \texttt{[PLH]} tokens may be inserted between the tokens from the same constraint. To prevent this from happening and to keep each constraint intact, we optionally prohibit inserting \texttt{[PLH]} within a multi-token constraint by constraining 0 to the number of such placeholders.

\paragraph{} In Figure~\ref{fig:levt_constraint}, our constraint insertion is executed at the first pass, and subsequent iterations start from deletion (indicated by a loop in the figure). We note that this step happens only at inference; during training, the original LevT training routine is carried out without the constraint insertion.

\section{Experiments}
\label{sec:exp}

We extend the \textsc{fairseq}\footnote{\href{https://github.com/pytorch/fairseq/commit/2d51e04}{https://github.com/pytorch/fairseq/commit/2d51e04}} \cite{ott2019fairseq} implementation of the original LevT architecture

to perform lexically-constrained decoding. All Transformer blocks in our LevT model follow the base configuration that contains 6 layers with 8 attention heads each, with a model size $d_\mathrm{model} = 512$ and feed-forward layer size $d_\mathrm{ff} = 2048$; the source and target embeddings share the same vocabulary. The LevT model is trained using knowledge distillation routine using Transformer base output released by \citet{gu2019levenshtein}. We leave more experimental details in the Appendix.

\subsection{Data and evaluation settings}

We evaluate our approach on the WMT'14 English-German (En-De) 
news translation task \citep{bojar-etal-2014-findings} with En-De bilingual dictionary entries extracted from Wiktionary\footnote{\href{https://dumps.wikimedia.org/enwiktionary/}{https://dumps.wikimedia.org/enwiktionary/}} following \citet{dinu-etal-2019-training}, by matching the source and target phrases of the dictionary entries in the source and target sentences, respectively.

We also evaluate our approach on two En-De test sets released by \citet{dinu-etal-2019-training} to compare our approach against previous work on applying lexical constraints in NMT \citep{post-vilar-2018-fast,dinu-etal-2019-training}. The two test sets are subsets of WMT'17 En-De test set \cite{bojar-etal-2017-findings} extracted using Wiktionary and the Interactive Terminology for Europe (IATE) terminology database,\footnote{\href{https://iate.europa.eu/}{https://iate.europa.eu/}} respectively. Both the WMT'14 and WMT'17 En-De datasets are tokenized using the Moses tokenization scripts and segmented into sub-word units using byte-pair encoding \cite{sennrich-etal-2016-neural}.

\subsection{Results}

We evaluate the systems using BLEU scores \cite{papineni-etal-2002-bleu} and \textit{term usage rate} (Term\%), which is defined as the number of constraints generated in the output divided by the total number of the given constraints.

\begin{table}
    \centering
    \small
    \begin{tabular}{l|cccc}
         \hline
         & \multirow{2}{*}{Term\%} & \multicolumn{2}{c}{BLEU} & Speed \\
         & & Full & Constr. & (sent/sec) \\
         \hline
         Baseline LevT               & 80.23 & 26.49 & 29.86 & 263.11 \\
         ~+ \textit{Constr. Ins.} & 94.43 & 26.50 & 29.93 & 260.19 \\
         ~~~+ \textit{No Del. }   & 99.62 & 26.59 & 30.43 & 260.61 \\
         ~~~~~+ \textit{No Ins. }   & \textbf{100.00} & \textbf{26.60} & \textbf{30.49} & 254.64 \\
         \hline
    \end{tabular}
    \caption{Results of LevT with lexical constraints on WMT14 En-De task}
    \label{tab:result_main}

\end{table}

\begin{table*}[ht]
    \centering
    \small
    \begin{tabular}{l|l}
    \hline
    Source & ``We don't want to \textbf{charge} that,'' she said. \\ \hline
    Baseline LevT & ``Das wollen wir nicht in Rechnung stellen'', sagte sie. \\
    ~+ \textit{Constr. Ins.} & ``Das wollen wir nicht verlangen'', sagte sie. \\
    ~~~+ \textit{No Del.} + \textit{No Ins.} & ``Das wollen wir nicht \textbf{berechnen}'', sagte sie. \\
    Reference & ``Wir möchten diese Summe nicht \textbf{berechnen}'', erklärte sie. \\
    \hline
    \end{tabular}
    \caption{Example translations from the LevT with constraint insertion to enforce the translation of \textit{charge}$\rightarrow$\textit{berechnen}. When deletion is allowed (+ \textit{Constr. Ins.}) the imposed constraint (\textit{berechnen}) gets deleted during decoding. But when deletion is disallowed (+ \textit{No Del.}) and unwanted insertion between constraint tokens is prohibited (+ \textit{No Ins.}), it guarantees the presence of our desired term in the final translation. We show more examples in the Appendix.}
    \label{tab:samples}
\end{table*}

Table~\ref{tab:result_main} shows the result of (i) the baseline LevT model, (ii) with the constraint insertion operation (\textit{+ Constr. Ins.}), (iii) with the constraint insertion operation and forcefully disallowing deletion of the constraints (\textit{+ No Del.}) and (iv) disallowing \texttt{[PLH]} insertion between tokens from the same constraint (\textit{+ No Ins.}). Table~\ref{tab:samples} shows an example where prohibiting constraint deletion prevents catastrophic removal of the lexical constraint.

We report results on both the filtered test set for sentence pairs that contain at least one target constraint (``Constr.'', 454 sentences) and the full test set (``Full'', 3,003 sentences).  
The constraint insertion operation increases the term usage rate from about 80\% to over 94\%, and further disallowing deletion of the constraints achieves above 99\% term usage. Prohibiting insertion between each constraint's tokens guarantees a 100\% term usage. For sentences with lexical constraints, we observe a statistically significant improvement of 0.6 BLEU ($p$-value $<$ 0.05) based on bootstrap resampling \citep{koehn-2004-statistical}. On the full test set, the BLEU improves by 0.1. The small margin of improvement is because only 1\% of the total reference tokens are constraint tokens. Unlike previous work that sacrificed decoding speed to enforce lexical constraints \citep[e.g.][]{hasler-etal-2018-neural,post-vilar-2018-fast}, there is no significant difference in the number of sentences decoded per second between the unconstrained and the lexically constrained LevT models. 

Table~\ref{tab:result_dinu} presents the comparison to two previous approaches: constrained decoding with dynamic beam allocation \cite{post-vilar-2018-fast} and data augmentation by replacing the source terms with target constraints during training \cite{dinu-etal-2019-training}. We refer to them as \textsc{Post18} and \textsc{Dinu19}, respectively, in Table~\ref{tab:result_dinu}. We evaluate each approach on the WMT'17 En-De test set with constraint terms from Wiktionary and IATE dictionaries.
Note that our baseline LevT model with Transformer blocks of 6 layers is superior to that of \citet{dinu-etal-2019-training} who used a 2-layer configuration. Despite having a stronger baseline, we obtain higher absolute BLEU score improvements (0.96 and 1.16 BLEU on Wiktionary and IATE, respectively) and achieved 100\% term usage. We report additional experiments on WMT'16 Romanian-English news translation task \cite{bojar-etal-2016-findings} in the Appendix.

\begin{table}
    \centering
    \small
    \begin{tabular}{l|cc|cc}
    \hline
    & \multicolumn{2}{|c}{Wiktionary} & \multicolumn{2}{|c}{IATE} \\
    & Term\% & BLEU & Term\% & BLEU \\
    \hline
    \multicolumn{5}{l}{\textit{Previous work}} \\
    \hline
    Baseline Trans. & 76.90 & 26.00 & 76.30 & 25.80 \\
    \textsc{Post18}    & 99.50 & 25.80 & 82.00 & 25.30 \\
    \textsc{Dinu19}   & 93.40 & 26.30 & 94.50 & 26.00 \\
    \hline
    \multicolumn{5}{l}{\textit{This work}} \\
    \hline
    Baseline LevT                         & 81.11 & 30.24 & 80.31 & 28.97 \\
    ~+ \textit{Constr. Ins.}  & 93.44 & 30.82 & 93.81 & 29.73 \\
    ~~~+ \textit{No Del.}       & 98.53 & 31.04 & 99.12 & 30.09 \\
    ~~~~~+ \textit{No Ins.}       & \textbf{100.00} & \textbf{31.20 }& \textbf{100.00} & \textbf{30.13} \\
    \hline
    \end{tabular}
    \caption{Comparison to previous work. Baseline Transformer and \textsc{Post18} results are from \citet{dinu-etal-2019-training}.}
    \label{tab:result_dinu}
\end{table}

\subsection{Analysis}

To analyze if our approach inserts the constraints at correct positions, we compare it to a baseline approach of randomly inserting the constraint terms in the output of our baseline LevT model. Note that we only insert those constraints that are not already present in the output. Although this results in a 100\% term usage, we observe that the BLEU score drops from 29.9 to 29.3 on the ``Constr.'' WMT'14 test set, whereas our approach improves the BLEU score. The LevT model with our proposed constraint insertion seems to inherently have the ability to place the constraints at correct positions in the target sentence.

Although prohibiting constraint deletion improves term usage in the final translation and achieves higher BLEU scores, it limits the possibility of reordering when there is more than one constraint during inference. For the English-German test sets we evaluated on, 97-99\% of the target constraints appear in the same order as the source terms. This issue may become more apparent in language pairs with more distinct syntactic differences between the source and target languages. In practice, most of the entries in terminology databases (Wiktionary, IATE, etc.) are often nominal. Thus, the reordering of lexical constraints boils down to whether the source and target language share the same argument-predicate order.\footnote{i.e., Subject (S), Verb (V), and Object (O), SOV, or VSO ordering.} We will explore potential strategies to reorder constraints dynamically in future work.

\section{Conclusion}

We proposed a non-autoregressive decoding approach to integrate lexical constraints for NMT. Our constraint insertion step is simple and we have empirically validated its effectiveness. The approach demonstrated control over constraint terms in target translations while being able to decode as fast as a baseline Levenshtein Transformer model, which achieves significantly higher decoding speed than traditional beam search.\footnote{Our implementation will be made publicly available at \href{https://github.com/raymondhs/constrained-levt}{https://github.com/raymondhs/constrained-levt}.} 
In addition to the terminological lexical constraints discussed in this work, future work can potentially modify insertion or selection operations to handle target translations of multiple forms; this can potentially disambiguate the morphological variants of the lexical constraints.

\bibliography{acl2020}
\bibliographystyle{acl_natbib}

\appendix

\section{Datasets}

We train on 3,961,179 distilled sentence pairs released by \citet{gu2019levenshtein} and evaluate on WMT'14 En-De test set (3,003 sentences). The dictionary used in this work is created by sampling 10\% En-De translation entries from Wiktionary, resulting in 10,522 entries. After applying this dictionary to generate constraints for the test set, we obtain 454 sentences that contain at least one constraint. The average number of constraints per sentence is 1.15 and the number of unique source constraints is 220. We use an English frequency list\footnote{\url{https://norvig.com/ngrams/count_1w.txt}} to filter the 500 most frequent words. We use the WMT'17 En-De test sets released by \citet{dinu-etal-2019-training}\footnote{\url{https://github.com/mtresearcher/terminology_dataset}} that were created based on Wiktionary and IATE term entries exactly matching the source and target. They contain 727 and 414 sentences, respectively.

\section{Hyperparameters}
\label{sec:hyper}

Table~\ref{tab:model} shows the hyperparameter settings for our LevT model. We learn a joint BPE vocabulary with 32,000 operations. Their resulting vocabulary size is 39,843.

\begin{table}[!ht]
    \centering
    \begin{tabular}{l|r}
    Embedding dim.           & 512 \\
    Learned positional embeddings & Yes \\
    Tied embeddings          & Yes \\
    Transformer FFN dim.     & 2,048 \\
    Attention heads          & 8 \\
    En/Decoder layers        & 6 \\
    Label smoothing          & 0.1 \\
    Dropout                  & 0.3 \\
    Weight decay             & 0.01 \\
    Learning rate            & 0.005 \\
    Warmup updates           & 10,000 \\
    Effective batch size in tokens     & 64,000 \\
    Max. updates              & 300,000 \\
    \hline
    \end{tabular}
    \caption{LevT hyperparameter settings}
    \label{tab:model}
\end{table}

\section{Additional Experiments}

We train a LevT model on 599,907 training sentence pairs from the WMT'16 Romanian-English (Ro-En) news translation task \cite{bojar-etal-2016-findings} using knowledge distillation routine based on Transformer base output and evaluate on 1,999 test sentences. Similar to En-De, we create a dictionary by sampling 10\% Ro-En translation entries from Wiktionary, resulting in 3,490 entries. We use this dictionary to generate 270 test sentences that contain at least one constraint. The average number of constraints per sentence is 1.11, and the number of unique source constraints is 122. Similarly, we filter out the 500 most frequent English words. 

We train our LevT model using the same hyperparameter settings from Table~\ref{tab:model}. We learn a joint BPE vocabulary with 40,000 operations, which results in 39,348 vocabulary size. Table~\ref{tab:result_add} shows the experiment results. We observe consistent findings in our En-De experiments in terms of improved term usage rate (from 80\% to 100\%) and a small margin of improvement of 0.7 BLEU, while being able to decode as fast as a baseline LevT model.

\begin{table}
    \centering
    \small
    \begin{tabular}{l|cccc}
         \hline
         & \multirow{2}{*}{Term\%} & \multicolumn{2}{c}{BLEU} & Speed \\
         & & Full & Constr. & (sent/sec) \\
         \hline
        Baseline LevT & 80.33 & 33.00 & 35.35 & 271.32 \\
        ~+ \textit{Constr. Ins.} & 95.33 & 33.10 & 35.96 & 274.01 \\
        ~~~+ \textit{No Del. } & 98.67 & \textbf{33.13} & \textbf{36.09} & 263.68 \\
        ~~~~~+ \textit{No Ins. }   & \textbf{100.00} & \textbf{33.13} & \textbf{36.09} & 264.45 \\
        \hline
    \end{tabular}
    \caption{Results of LevT with lexical constraints on WMT16 Ro-En task}
    \label{tab:result_add}

\end{table}

\section{Examples}

Table~\ref{tab:samples_more} shows more example translations of the lexically constrained LevT model.

\begin{table*}[ht]
    \centering
    \scriptsize
    \begin{tabular}{l|l}
\multicolumn{2}{l}{\textit{WMT'14 En-De}} \\
\hline
Source & Bwelle and his team \textbf{spend almost} every weekend seeing hundreds of patients \{\textit{spend}$\rightarrow$\textit{verbringen}, \textit{almost}$\rightarrow$\textit{beinahe}\} \\
Baseline LevT & Bwelle und sein Team \textbf{verbringen} fast jedes Wochenende mit Hunderte von Patienten. \\
~+ \textit{Constr. Ins.} & Bwelle und sein Team \textbf{verbringen beinahe} jedes Wochenende mit Hunderte von Patienten. \\
~~~+ \textit{No Del.} + \textit{No Ins.} & Bwelle und sein Team \textbf{verbringen beinahe} jedes Wochenende mit Hunderte von Patienten. \\
Reference & Bwelle und sein Team \textbf{verbringen beinahe} jedes Wochenende damit, Hunderte von Patienten zu behandeln \\
\hline
Source & There have already been two events held in the brightly lit café. \{\textit{already}$\rightarrow$\textit{schon}\} \\
Baseline LevT & Im hell beleuchteten Café fanden bereits zwei Veranstaltungen statt. \\
~+ \textit{Constr. Ins.} & Im hell beleuchteten Café fanden bereits zwei Veranstaltungen statt. \\
~~~+ \textit{No Del.} + \textit{No Ins.} & Im hell beleuchteten Café fanden \textbf{schon} zwei Veranstaltungen statt. \\
Reference & Zwei Events gab's auch \textbf{schon} im hellen Café. \\
    \hline
\multicolumn{2}{l}{\textit{WMT'17 En-De - Wiktionary}} \\
\hline
Source &  House searches had revealed \textbf{evidence} and drugs, the \textbf{police} revealed on Friday. \{\textit{evidence}$\rightarrow$\textit{Beweismittel}, \textit{police}$\rightarrow$\textit{Polizei}\} \\
Baseline LevT & Durchsuchungen des Hauses hatten Beweise und Drogen enthüllt, die \textbf{Polizei} am Freitag enthüllt. \\
~+ \textit{Constr. Ins.} & Hausdurchfragen hatten \textbf{Beweismittel} und Drogen offenbart, hat die \textbf{Polizei} am Freitag enthüllt. \\
~~~+ \textit{No Del.} + \textit{No Ins.} & Durchfragen hatten \textbf{Beweismittel} und Drogen offenbart, die \textbf{Polizei} am Freitag enthüllt. \\
Reference & Bei Wohnungsdurchsuchungen seien \textbf{Beweismittel} und Rauschgift sichergestellt worden, teilte die \textbf{Polizei} am Freitag mit. \\
\hline
Source & We always say that it has a lot of \textbf{Latin American} influences. \{\textit{Latin American}$\rightarrow$\textit{lateinamerikanisch}\} \\
Baseline LevT & Wir sagen immer, dass sie viele lateinamerikanische Einflüsse hat. \\
~+ \textit{Constr. Ins.} & Wir sagen immer, dass sie viel \textbf{lateinamerikanisch} beeinflusst. \\
~~~+ \textit{No Del.} + \textit{No Ins.} & Wir sagen immer, dass sie viel \textbf{lateinamerikanisch} beeinflusst. \\
Reference & Wir sagen immer, dass sie sehr \textbf{lateinamerikanisch} geprägt ist. \\
\hline
\multicolumn{2}{l}{\textit{WMT'17 En-De - IATE}} \\
\hline
Source &  What is behind \textbf{sleep disorders}? \{\textit{sleep disorders}$\rightarrow$\textit{Schlafstörungen}\} \\
Baseline LevT & Was steckt hinter Schlafkrankheiten? \\
~+ \textit{Constr. Ins.} & Was steckt hinter \textbf{Schlafstörungen}? \\
~~~+ \textit{No Del.} + \textit{No Ins.} & Was steckt hinter \textbf{Schlafstörungen}? \\
Reference & Was steckt hinter \textbf{Schlafstörungen}? \\
\hline
Source &  He said another stepson who lives nearby alerted him. \{\textit{stepson}$\rightarrow$\textit{Stiefsohn}\} \\
Baseline LevT &  Er sagte, ein weiterer Stiefson, der in der Nähe lebt, alarmierte ihn.\\
~+ \textit{Constr. Ins.} & Er sagte, ein weiterer \textbf{Stiefsohn}, der in der Nähe lebt, alarmierte ihn. \\
~~~+ \textit{No Del.} + \textit{No Ins.} & Er sagte, ein weiterer \textbf{Stiefsohn}, der in der Nähe lebt, alarmierte ihn. \\
Reference & Er sagte, dass ihn ein weiterer \textbf{Stiefsohn}, der in der Nähe wohnt, gewarnt hätte. \\
\hline
    \end{tabular}
    \caption{More example translations from the LevT with constraint insertion. The constraints are in curly brackets.}
    \label{tab:samples_more}
\end{table*}

\end{document}